\documentclass[letterpaper, 10 pt, conference]{ieeeconf}  

\IEEEoverridecommandlockouts                              

\usepackage{graphics} 
\usepackage{epsfig} 
\usepackage{empheq}
\usepackage{times} 
\usepackage{amsmath} 
\usepackage{amssymb}  
\usepackage{bm}
\usepackage{color}
\usepackage{bbding} 
\usepackage{cite}
\usepackage{diagbox}
\usepackage[linesnumbered,ruled]{algorithm2e}
\usepackage{ulem} 
\usepackage{hyperref}
\usepackage{float}
\usepackage{booktabs}
\usepackage{multirow}
\usepackage{mathrsfs}
\usepackage[utf8]{inputenc}
\usepackage{subfigure}
\usepackage{graphicx}
\usepackage{pifont}
\usepackage{threeparttable}
\newcounter{RNum}

\newcommand{\etal}{{et al}. }
\newcommand{\ie}{{i}.{e}., }
\newcommand{\eg}{{e}.{g}., }
\newcommand{\etc}{{etc}.}
\makeatletter
\newcommand{\rmnum}[1]{\romannumeral #1}
\newcommand{\Rmnum}[1]{\expandafter\@slowromancap\romannumeral #1@}
\makeatother
\usepackage[T1]{fontenc}
\usepackage{aecompl}
\usepackage{units}
\hypersetup{
	colorlinks,
	urlcolor=red,      
}

\begin{document}

	\title{\LARGE \bf
		DCPT: Darkness Clue-Prompted Tracking in Nighttime UAVs
	}

	\author{Jiawen Zhu$^{1\dagger}$, Huayi Tang$^{1\dagger}$, Zhi-Qi Cheng$^{2}$, Jun-Yan He$^{3}$, \\
		Bin Luo$^{3}$, Shihao Qiu$^{1}$, Shengming Li$^{1,*}$,  Huchuan Lu$^{1}$ 
		\thanks{$^{\dagger}$Equal contribution, $^{*}$Corresponding author}
		\thanks{$^{1}$Jiawen Zhu, Huayi Tang, Shihao Qiu, Shengming Li,  and Huchuan Lu are with Dalian University of Technology, Dalian 116024, China. {\tt\footnotesize lishengming@dlut.edu.cn}}
		\thanks{$^{2}$Zhi-Qi Cheng is with Carnegie Mellon University, PA 15213, USA.}
		\thanks{$^{3}$Jun-Yan He and Bin Luo are with DAMO Academy, Alibaba Group, Shenzhen 518000, China.}
	}

	\maketitle
	\thispagestyle{empty}
	\pagestyle{empty}

\begin{abstract}
	Existing nighttime unmanned aerial vehicle (UAV) trackers follow an “Enhance-then-Track” architecture - first using a light enhancer to brighten the nighttime video, then employing a daytime tracker to locate the object. 
	This separate enhancement and tracking fails to build an end-to-end trainable vision system. 
	To address this, we propose a novel architecture called Darkness Clue-Prompted Tracking (DCPT) that achieves robust UAV tracking at night by efficiently learning to generate darkness clue prompts.
	Without a separate enhancer, DCPT directly encodes anti-dark capabilities into prompts using a darkness clue prompter (DCP).
	Specifically, DCP iteratively learns emphasizing and undermining projections for darkness clues. It then injects these learned visual prompts into a daytime tracker with fixed parameters across transformer layers.
	Moreover, a gated feature aggregation mechanism enables adaptive fusion between prompts and between prompts and the base model.
	Extensive experiments show state-of-the-art performance for DCPT on multiple dark scenario benchmarks.
	The unified end-to-end learning of enhancement and tracking in DCPT enables a more trainable system. The darkness clue prompting efficiently injects anti-dark knowledge without extra modules.
	Code is available at \url{https://github.com/bearyi26/DCPT}.
\end{abstract}
	
\section{Introduction}
Visual object tracking from unmanned aerial vehicles (UAVs) is an essential capability of aerial robotic vision, enabling various downstream applications such as traffic monitoring~\cite{tian2011video}, aerial cinematography~\cite{bonatti2019towards}, and search and rescue~\cite{al2019appearance}. 
While recent advances using deep neural networks~\cite{alexnet, resnet, vit} and large-scale datasets~\cite{uav, got10k, trackingnet} have achieved promising tracking performance in daytime conditions, state-of-the-art trackers~\cite{siamapn++, transt, ostrack} still struggle in more challenging nighttime environments.
When faced with more challenging light conditions (\eg the night falls), these approaches often suffer from severe performance degradation or even fail to work.
This is mainly because discriminative visual cues like color and geometry are diminished at night, and onboard cameras introduce more noise and image degradation under low illumination. As a result, existing trackers fail to extract robust features for accurate target localization at night. 
Therefore, to fully realize the potential of UAV vision, it is imperative to explore effective night-time tracking techniques, which will promote the versatility and survivability of UAV vision systems.

\begin{figure}[!t]
	\centering
	\setlength{\abovecaptionskip}{-0.0cm}
	\vspace{-0.17cm}
	\includegraphics[width=0.905\columnwidth]{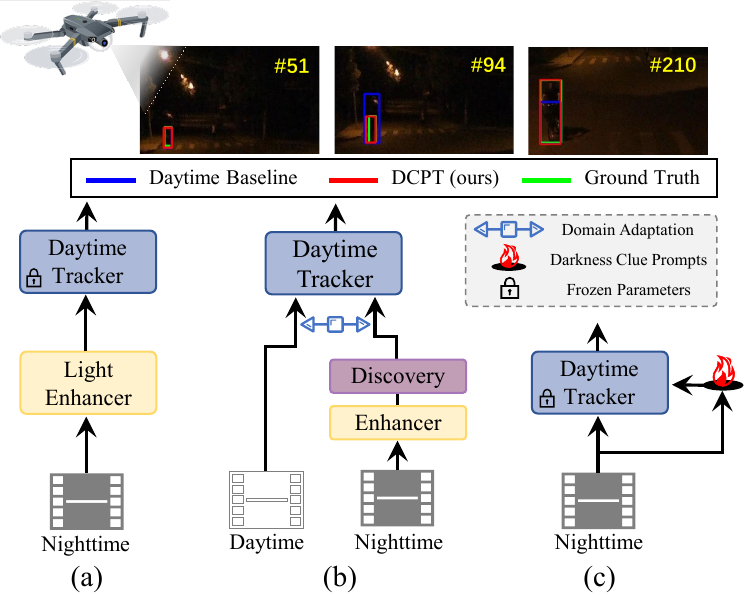}
	\vspace{-0.1cm}
	\caption{Illustration of different nighttime UAV tracking paradigms.
		(a) “Enhance-then-Track" paradigm. 
		(b) Domain adaptation paradigm.
		(c) Proposed darkness clue-prompted tracking (DCPT) paradigm.
		DCPT possesses a more streamlined structure while effectively incorporating the learned darkness clue prompts, enabling the UAV to “see” sharper in the dark.
	}
	\label{fig:intro}
	\vspace{-0.1cm}
\end{figure}

Several studies have been conducted to equip UAV systems with the capacity to "see" in low-light conditions.
For example, Fu \etal \cite{highlightnet} introduced a light enhancer called "HighlightNet" designed to illuminate specific target areas for UAV trackers.
This “Enhance-then-Track" paradigm (Fig.~\ref{fig:intro} (a)) is also adopted in other studies~\cite{darklighter,adtrack}.
They mainly focus on designing a light enhancer and employ off-the-shelf daytime trackers to output the final tracking results.
On the other hand, Ye \etal~\cite{nat2021} introduce domain adaptation (Fig.~\ref{fig:intro} (b)) for nighttime UAV tracking.
They generate nighttime training samples and adversarially train a model for narrowing the gap between day and night circumstances.
Despite gaining improvements, current solutions for nighttime tracking still have significant limitations.
\textbf{\rmnum{1})} Performing image enhancement before tracking makes UAV vision system over-reliance on the extra trained enhancer, and separating the nighttime tracking into two sub-processes masks is not conducive to building an end-to-end trainable architecture. 
\textbf{{\rmnum{2})}} Domain adaptation requires abundant data for training and high-quality target domain samples are scarce for nighttime domain learning. 
\textbf{{\rmnum{3})}} The intrinsic relationship between daytime tracker and nighttime tracker is overlooked, and the potential for employing daytime tracker in nighttime scenarios is not well exploited.

Recently, prompt learning has attracted much attention, extending from Natural Language Processing (NLP) to the vision tasks~\cite{doprompt,vpt,vipt}. 
Typically, the foundation model is frozen, and only a few tunable parameters are added for learning valid prompts. This approach demonstrates promising results and efficiencies.
Drawing inspiration from these works, we formulate nighttime UAV tracking as a prompt learning problem where the goal is to mine valid darkness clue prompts for a well-trained daytime tracker, so that the parameter and computational cost are constrained and the nighttime performance is maximized.
In this work, we propose a novel nighttime UAV tracker (termed DCPT) that dwells on learning darkness clue prompting for low-light circumstances (Fig.~\ref{fig:intro} (c)).
Specifically, to effectively facilitate the discovery and mining of clues in darkness, we design a darkness clues prompter (DCP, in \ref{sec:dcp}) by introducing the back-projection structure which shows favorable performance in image restoration (\eg image super-resolution~\cite{dbpn}). 
DCP propagates the darkness clue prompts across all semantic layers of the foundation tracker.
Moreover, we design a gated feature aggregation (GFA, in \ref{sec:gfa}) mechanism to efficiently fuse bottom-up prompts and enable the complementary integration of learned prompts and information from the foundation model.
Ultimately, DCPT can effectively inject learned darkness clue prompts into a frozen daytime model with only a small number of prompt learning-related parameters, obtaining superior nighttime tracking performance.

We summarize our major contributions as follows:
\begin{itemize}
	\item We propose DCPT, a novel solution for nighttime UAV tracking by introducing darkness clue prompt learning. In this way, the tracker's potential is stimulated by the learned prompts in extreme low-light circumstances.
	
	\item Darkness clue prompter is proposed for mining valid visual prompts at night. Besides, a gated feature aggregation mechanism is designed for effectively fusing the features between prompters and the foundation model.
	
	\item Extensive experiments on four nighttime tracking benchmarks validated the effectiveness of DCPT. Qualitative and quantitative results demonstrate the superiority of DCPT as a nighttime tracker for aerial robots.
	\eg on DarkTrack2021~\cite{darktrack}, DCPT boosts the base tracker by 4.9\% success score with 3.0M prompting parameters.
\end{itemize}

\section{Related Works}
\subsection{Nighttime UAV Tracking}
The emergence of deep-learning technologies and efforts by researchers
push visual object tracking to new frontiers\cite{siameserpn,dimp,transt,ostrack}.
Impressive tracking performance was also achieved on the UAV platforms \cite{siamapn,siamapn++}.
However, these trackers, which are mainly designed for daytime scenarios, often suffer severe performance degradation or even fail to work when faced with common but challenging nighttime scenarios.
This is because the nighttime circumstances suffer from loss of detailed information, accompanying noise, and low contrast and brightness.
Therefore, nighttime UAV tracking has attracted increasing attention.
A straightforward manner is to complete nighttime tracking by an "Enhance-then-Track" process.
Specifically, researchers design low light enhancers~\cite{darktrack,darklighter,highlightnet} for 
nighttime scenarios, and use existing daytime UAV trackers to track in the enhanced sequences.
Li \etal~\cite{adtrack} propose to integrate a low-light image enhancer into a CF-based tracker for robust tracking at night. Similarly, DarkLighter~\cite{darklighter} and HighlightNet~\cite{highlightnet} also design low-light image enhancers to alleviate the influence of extreme illumination and highlight the potential objects, respectively.
Although effective, the main drawbacks of such approaches are that the nighttime tracker requires an additional trained light enhancer for preprocessing, incurring extra computational costs, and the separation of enhancer and tracker is not conducive to building an end-to-end trainable nighttime UAV tracking method.
Another approach is to adopt domain adaptation to transfer daytime tracker to nighttime scenarios.
To obtain the tracking capabilities in the dark, UDAT~\cite{nat2021} proposes to align image features from daytime and nighttime domains by the transformer-based bridging layer, in this way, the tracking capabilities on the daytime domain are somewhat transferred to the nighttime domain.
Unfortunately, this approach requires more training costs and the lack of high-quality target domain data for tracking also limits its enhancement.

\begin{figure*}[!t]
	\centering
	\includegraphics[trim=15 0 0 0, clip, width=0.925\textwidth]{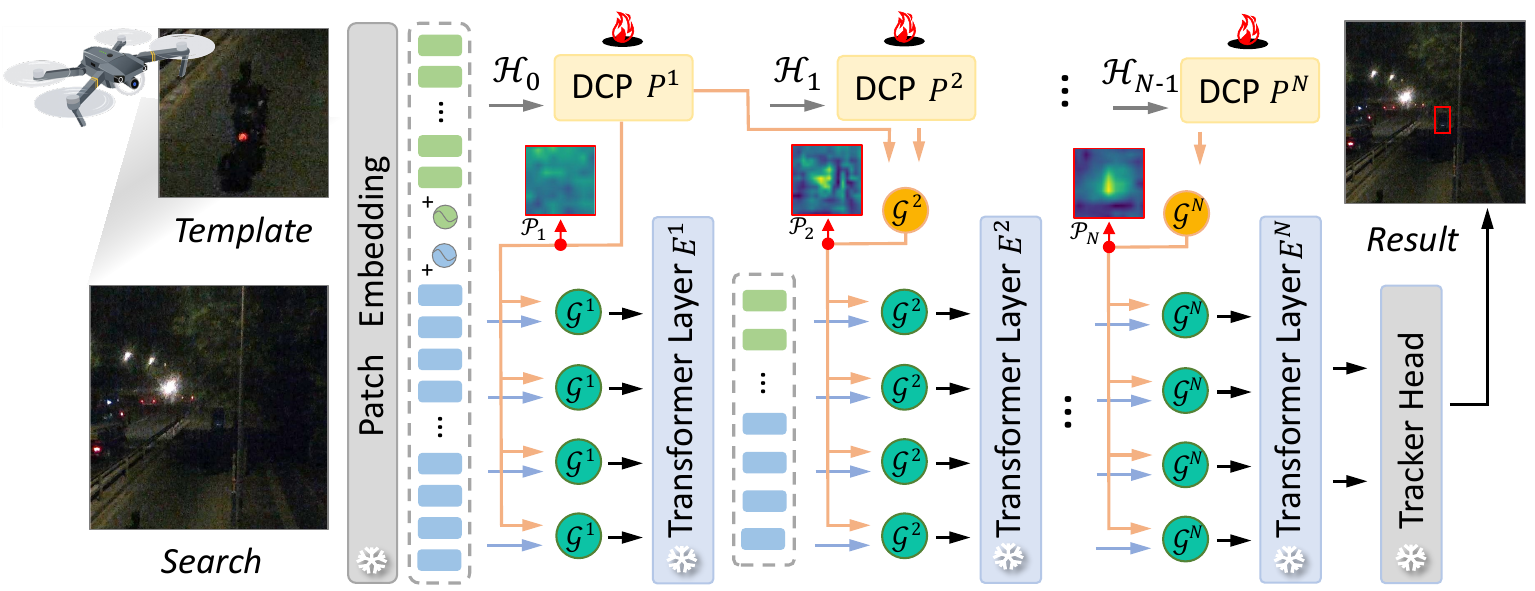}
	\vspace{-3mm}
	\caption{\textbf{Overview architecture of DCPT.} 
		The template and search images are first fed into the patch embedding to generate the corresponding tokens. 
		A ViT backbone 
		is employed for fundamental feature extraction and interaction of the concatenated template and search tokens. 
		In parallel, the darkness clue prompter (DCP) blocks $P^i, i\!\in\!\{1,...,N\}$ are distributed in each encoder layer,
		and they are responsible for extracting valid darkness clue prompts and injecting them into the foundation model. Besides, the gated feature aggregation (GFA) is performed for more effective information fusion. 
	}
	\vspace{-3mm}
	\label{fig:overview}
\end{figure*}

\subsection{Visual Prompt Learning}
In natural language processing (NLP), a pre-trained model can easily adapt to its downstream tasks by incorporating specific prompts to the input text.
As a parameter-efficient learning manner, prompt tuning begins to make its mark on visual tasks~\cite{vpt,bahng2022visual,vipt}.
VPT~\cite{vpt} is among the pioneers in exploring visual prompt tuning. It adds a small number of learnable parameters to the pre-trained foundation model, and obtains promising results compared with full fine-tuning on multiple downstream tasks. 
Instead of focusing on embedding space, Bahng \etal ~\cite{bahng2022visual} propose to train learnable perturbations in the embedding pixel space.
For multi-modal tracking, ViPT~\cite{vipt} proposes to learn efficient auxiliary-modal prompts for foundation RGB tracker, achieving impressive multi-modal tracking performance.
The success of these methods above shows us the potential of prompt learning for nighttime UAV tracking.
However, unlike multi-modal tracking, which has an additional auxiliary-modal flow that can be used directly to learn prompts, nighttime UAV tracking only has extreme lighting scenarios as input.
In this work, to overcome the UAV tracker's poor performance in indistinguishable night scenarios,
we propose to mine the darkness clues as effective visual prompts for a daytime tracker, enabling a sharper vision in the dark.

\section{Methodology}
\subsection{Overview}
The overall architecture is shown in Fig.~\ref{fig:overview}.
To summarize, DCPT injects the learned darkness clue prompts $\mathcal{P}^{i}, i\!\in\!\{1,...,N\}$ into the daytime tracker and stimulates its tracking potential in low-light circumstances.
The darkness clue prompts are uncovered through end-to-end prompt learning in nighttime data, having the ability to discriminate object tracks in the darkness, which is used to complement the shortcomings of the daytime tracker.
These learned darkness clue prompts are propagated from preceding prompts and foundation feature flows $\mathcal{H}^{i}, i\!\in\!\{0,...,N-1\}$.
Specifically, the gated feature aggregation (GFA) mechanism $\mathcal{G}^{i}, i\!\in\!\{1,...,N\}$ is designed for controlling the weight of these different information sources. 
In general,
the daytime tracker has excellent tracking capabilities on generic scenarios while just lacking specific adaptations for nighttime scenarios.
Therefore, we only tune parameters that are related to darkness clue prompt generation instead of fine-tuning the entire model.
The DCPT framework
maximizes the capabilities inherited from daytime tracker trained on large-scale datasets~\cite{got10k,lasot,trackingnet}, avoiding overfitting on limited nighttime tracking data but gaining nighttime-specific tracking capabilities through darkness clue prompt learning.

\subsection{Daytime Foundation Tracker}
Daytime and nighttime trackers naturally share a number of basic capabilities, including scene understanding and feature matching between targets,
hence,  
previous nighttime tracking methods~\cite{adtrack,highlightnet,darklighter} all chose cutting-edge trackers~\cite{li2019siamrpn++,siamapn++,dimp} as their daytime base trackers.
Hereby 
we adopt a streamlined model to serve as a clear baseline, emphasizing the advantages of our darkness clue prompting tracking framework.
Specifically, 
the base tracker is only composed of two parts: 
a backbone and a prediction head.
The backbone consists of stacked vision transformer (ViT) layers and propagating the inputs in a one-stream style like the method in \cite{ostrack}.
First, both template $\bm{Z}\in\mathbb{R}^{H_{z}\times W_{z}\times 3}$ and search region $\bm{X}\in\mathbb{R}^{H_{x}\times W_{x}\times 3}$ are embedded and flattened to 1D tokens with added positional embedding ${E}_{pos}$:
\begin{align}
	\label{eq:vit_emb}
	\bm{\mathcal{H}}_Z, \bm{\mathcal{H}}_X = {E}_{embed}(\bm{Z};\bm{X})+{E}_{pos},
\end{align}
The template and search tokens are then concatenated 
to $\bm{\mathcal{H}}^{0}_{base}=concat(\bm{\mathcal{H}}_Z, \bm{\mathcal{H}}_X)$
and passed through a $N$-layer standard vision transformer encoder:
\begin{align}
	\label{eq:vit}
	\bm{\mathcal{H}}^l = {E}^{l}(\bm{\mathcal{H}}^{l-1}),  &&l=1, 2 \ldots, N
\end{align}
Last, we employ a lightweight corner head for box prediction.
The tracking results can be obtained by:
\begin{align}
	\label{eq:base_head}
	\bm{B} = \phi(\bm{\mathcal{\mathcal{H}}}^{N}),
\end{align}
The prediction head $\phi$ does not require any complex
post-processes (\eg cosine window and size penalty), without any hyper-parameters, keeping our foundation model concise.

\subsection{Darkness Clue Prompter}
\label{sec:dcp}
The foundation model is well-trained on daytime data and therefore does not have sufficient object-discriminating abilities in nighttime scenarios.
Therefore, we propose to equip the foundation model with the proposed darkness clue prompters (DCP), injecting the mined darkness clue prompts into the foundation feature flow.
The darkness clue prompting process can be formulated as:
\begin{align}
	\label{eq:prompt_overview}
	\bm{\mathcal{H}}_{p}^{l-1} = \bm{\mathcal{H}}^{l-1}+\bm{\mathcal{P}}^{l},
\end{align}
where $\bm{\mathcal{H}}_{p}^{l-1}$ is the prompted tokens which absorb the learned darkness clue prompt $\bm{\mathcal{H}}_{p}^{l}$ from the $l$-th DCP block.

\begin{figure}[t]
	\centering
	\includegraphics[trim=-12 0 0 0, clip, width=0.425\textwidth]{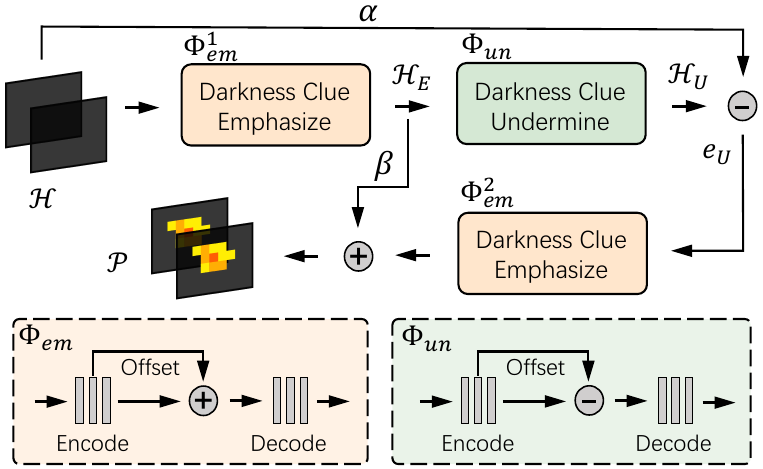}
	\vspace{-2mm}
	\caption{\textbf{Detailed structure of the proposed DCP module.} 
		DCP module takes the foundation features  
		as the input, iteratively emphasizing and undermining the darkness clues, learning the residual term for the reconstruction of valid darkness clue prompts. 
	}
	\label{fig:prompt}
	\vspace{-2mm}
\end{figure}

For nighttime UAV tracking, given low-light inputs without explicit learning objectives of darkness clue region, it is difficult to learn valid darkness clue prompts.
To address this problem, we introduce the back-projection~\cite{irani1991improving} ideology from image super-resolution (SR) into darkness clue prompt learning.
Given a immediate SR image $\bm{\mathcal{I}}_{t}$ ($t$ denotes the iteration index), the SR image $\bm{\mathcal{I}}_{t+1}$ can be obtained by back-projection operation:
\begin{align}
	\label{eq:bp}
	\bm{\mathcal{I}}_{t+1} = \bm{\mathcal{I}}_{t}+\lambda \Phi_{up}(\bm{\mathcal{I}}_{0}-\Phi_{down}(\bm{\mathcal{I}}_{t})),
\end{align}
where $\Phi_{up}$ and $\Phi_{down}$ denote the up-sampling and down-sampling functions, respectively. $\bm{\mathcal{I}}_{0}$ represent the initial low-super-resolution (LR) image. $\lambda$ is a balance coefficient for residual updating. 
Instead of learning the SR image in a direct feed-forward manner,
back-projection block iteratively mines the reconstruction error between LR and down-sampled SR images then fuses it back to tune the HR image.
We think of the philosophy of SR image refining as analogous to our darkness clue prompting process.
In this work, we construct the darkness clue prompt learning through the following equations:
\begin{align}
	&\bm{\mathcal{P}} = \beta \bm{\mathcal{H}}_{E}+{\Phi^{2}_{em}}(\bm{\mathcal{H}}_{U} -\alpha\bm{\mathcal{H}}_{E}),
	\label{eq:prompt} \\
	&\bm{\mathcal{H}}_{E} = \Phi^{1}_{em}(\bm{\mathcal{H}}),\ \bm{\mathcal{H}}_{U} = \Phi_{un}({\Phi^{1}_{em}}(\bm{\mathcal{H}})),
\end{align}
where $\Phi_{em}$ and $\Phi_{un}$ denote the prompt emphasizing and undermining functions, respectively. $\alpha, \beta \in \mathbb{R}$ are weights to balance the residual clues. We omit layer index $l$ for better readability.
Fig.~\ref{fig:prompt} showcases the proposed DCP block in detail.
The input features $\bm{\mathcal{H}}$ (we omit the reshape operation for simplicity) go through the first darkness clue emphasize block 
and forms prompted $\bm{\mathcal{H}}_{E}$.
Next, $\bm{\mathcal{H}}_{U}$ is obtained through a darkness clue undermine block from the estimated $\bm{\mathcal{H}}_{E}$.
Then we get the residual $\bm{e}_{U}$ which represents the difference between estimated  $\bm{\mathcal{H}}_{U}$ and the original $\bm{\mathcal{H}}$.
After that, another darkness clue emphasize is employed to generate the residual between darkness clue prompts $\bm{\mathcal{P}}$ and the estimated $\bm{\mathcal{H}}_{E}$. 
Finally, darkness clue prompts $\bm{\mathcal{P}}$ is produced by adding the estimated residual.
Borrowed from \cite{wang2020lightening}, the darkness clue emphasize and undermine blocks that consist of an encode-decode structure with plus and minus offset operations, respectively.
The proposed DCP blocks are attached to each encoder layer of the foundation model, iteratively learning the construction of valid darkness clue prompts for nighttime circumstances.

\subsection{Gated Feature Aggregation}
\label{sec:gfa}
The encoder of the foundation model possesses different information hierarchies across the blocks, while the darkness clue prompter (DCP) blocks are placed independently in front of each encoder layer.
This poses challenges for effective and efficient darkness clue prompt learning from two perspectives:
\textbf{\rmnum{1})} The learned darkness clue prompts lack front-to-back semantic hierarchies like foundation features, and learning different prompt blocks independently is inefficient.
\textbf{\rmnum{2})} For different semantic hierarchies and different spatial locations, the region that the darkness clue prompts focus on should also differ.
Direct additive injection of prompts lacks the flexibility to adaptively adjust to different regions.
To this end, we propose a gated feature aggregation (GFA) for learned prompts, as well as prompts and the foundation features.  
As shown in Fig.~\ref{fig:overview}, we perform gated aggregation for current $\bm{\mathcal{P}}^l$ and the preceding one $\bm{\mathcal{P}}^{l-1}$. The gated aggregation for adjacent prompts can be formulated as:
\begin{align}
	\label{eq:gate_prom}
	&\bm{\mathcal{P}}_{g}^{l+1} = {g^{l+1}}\times\bm{\mathcal{P}}^{l+1}+(1-{g^{l+1}})\times\bm{\mathcal{P}}^{l}_{g},\\
	&{g}^{l}=\nicefrac{1}{1+e^{-\gamma^{l}}}, \qquad  \bm{\mathcal{P}}^{1}_{g}=\bm{\mathcal{P}}^{1},
\end{align}
where the gated weights $g$ are generated through a sigmoid function, controlled by learnable factors $\gamma$. 
The gated aggregation connects DCP blocks from shallow to deep, thus, darkness clue prompts can accomplish bottom-up propagation across different feature hierarchies. 
This promotes the efficient learning of valid prompts, in addition, it is also proved effective for self-supervised classification~\cite{yoo2023improving}.
Moreover, we design gated aggregation for learned darkness clue prompts and foundation features at a finer granularity. The final prompted foundation feature can be obtained by:
\begin{align}
	\label{eq:prompt_overview}
	&\bm{\mathcal{H}}_{p,g}^{l-1} = \bm{\mathcal{H}}^{l-1}+\bm{p}^{l}\times\bm{\mathcal{P}}^{l},\\
	&\bm{p}^{l} = [{p_1^{l},p_2^{l},\ldots,p_M^{l}}],\\
	&{p}^{l}_{i}=\nicefrac{1}{1+e^{-\gamma^{l}_{i}}},  \qquad  i=1,2, \ldots, M
\end{align}
where $M$ denotes the number of the learned prompt tokens.
The gated aggregation weights $\bm{p}\in \mathbb{R}^{1\times M}$ 
indicate that different attention weights are assigned for darkness clue prompts from different regions corresponding to different tokens.
The proposed gated feature aggregation mechanism introduces negligible number of parameters but effectively improves the learning of darkness clue prompt and achieves higher nighttime tracking performance.

\subsection{Training Objective}
For object locating, we combine the $\mathcal{L}_1$ loss and the GIoU loss~\cite{giou} $\mathcal{L}_G$,
which can be formulated as:
\begin{align}
	\label{eq:locate_loss}
	\mathcal{L}_{locate} &= \lambda_1\mathcal{L}_1(\bm B, \bm{B}_{gt}) + \lambda_G\mathcal{L}_G(\bm B, \bm{B}_{gt}),
\end{align} 
where $\bm{B}_{gt}$ represents the ground truth, $\lambda_1=5$ and $\lambda_G=2$ are the weight parameters.
The same training objective is adopted for training the foundation and nighttime trackers.

\section{Experiment}
\subsection{Implementation Details}

\textbf{Foundation Tracker Training.}
We train the daytime foundation tracker on four common datasets, including GOT-10k~\cite{got10k}, LaSOT~\cite{lasot}, TrackingNet~\cite{trackingnet}, and COCO~\cite{coco}. 
The backbone is initialized from pre-trained MAE~\cite{mae}.
The model is trained with AdamW optimizer~\cite{adamw} for 300 epochs with a total batch size of 128, each epoch involves 60,000 sampling pairs.
The template and search region size are set to 128$\times$128 and 256$\times$256, respectively.

\begin{figure*}[!t]	
	\raggedright
	\includegraphics[width=0.325\linewidth]{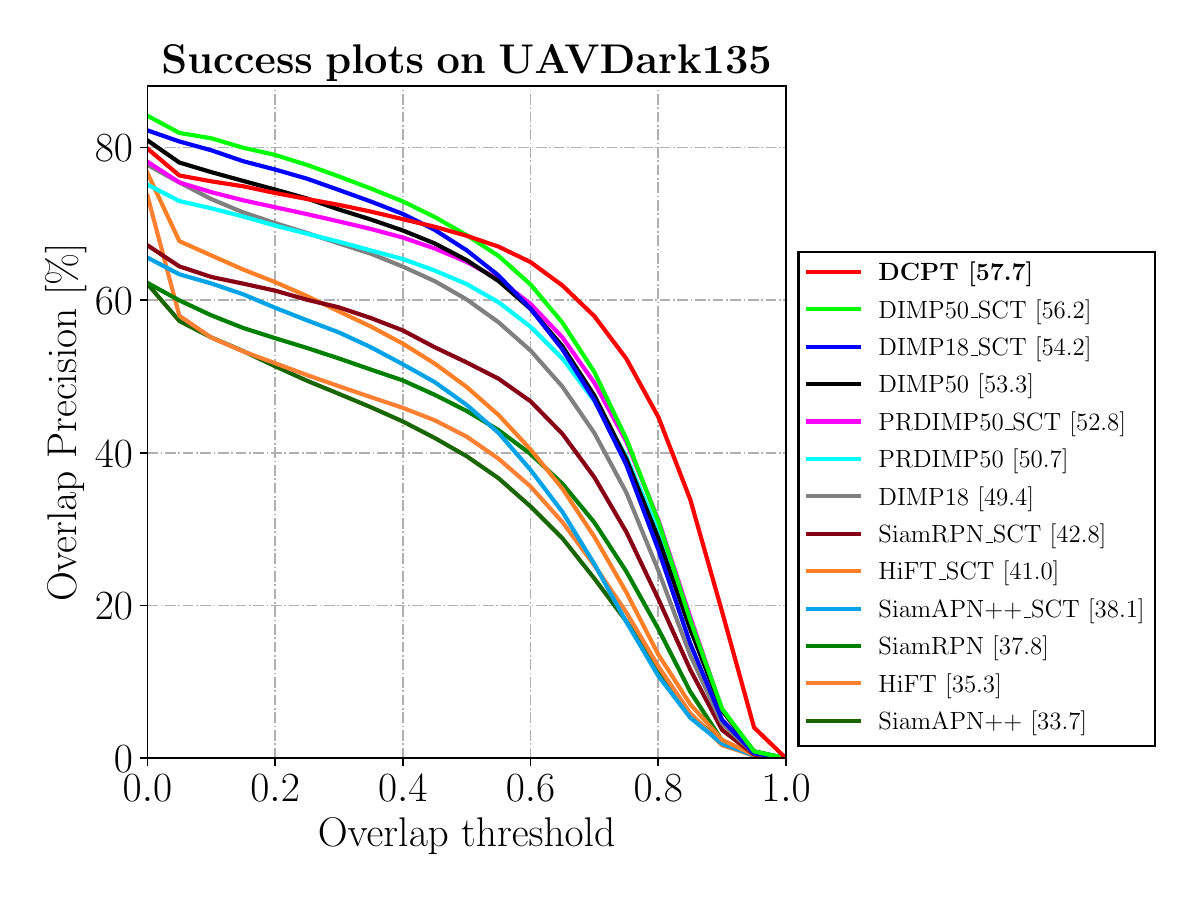}
	\includegraphics[width=0.325\linewidth]{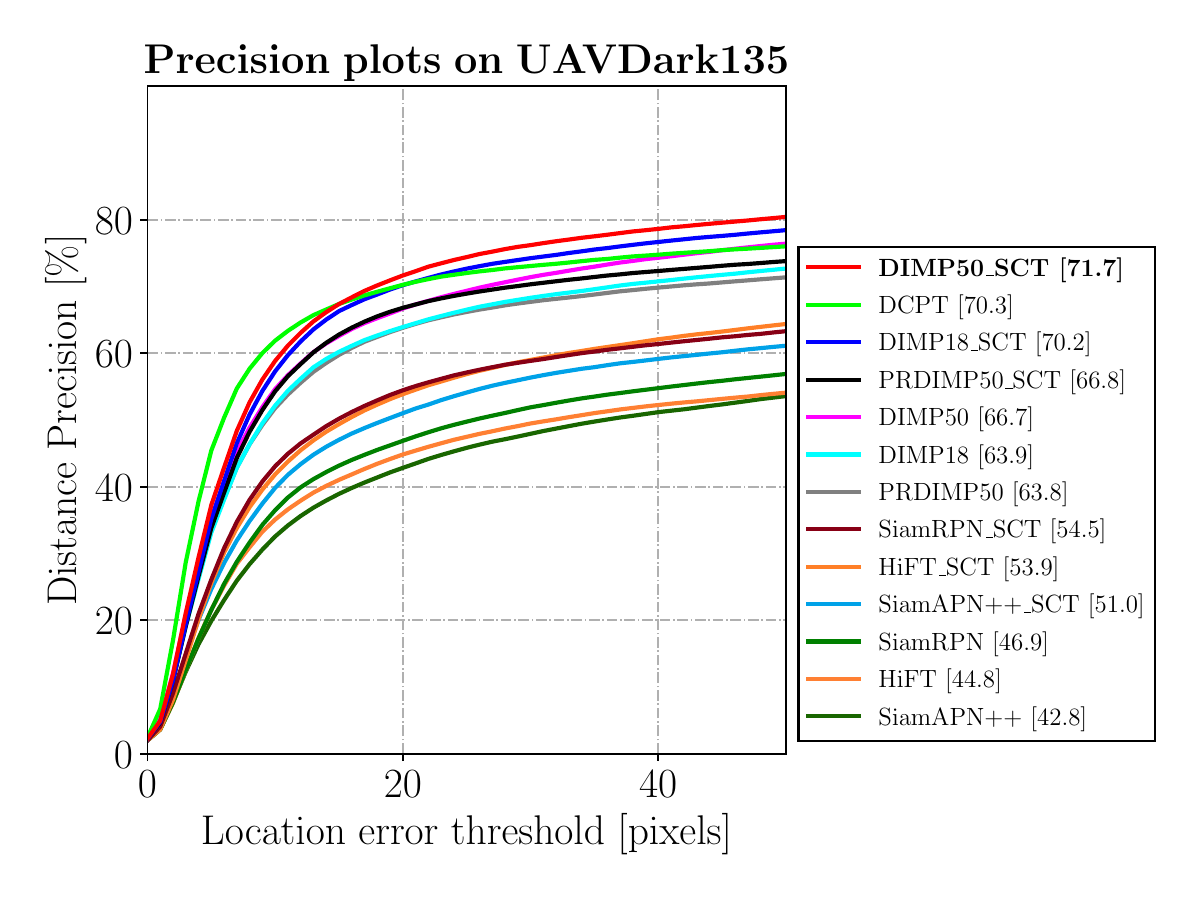}
	\includegraphics[width=0.325\linewidth]{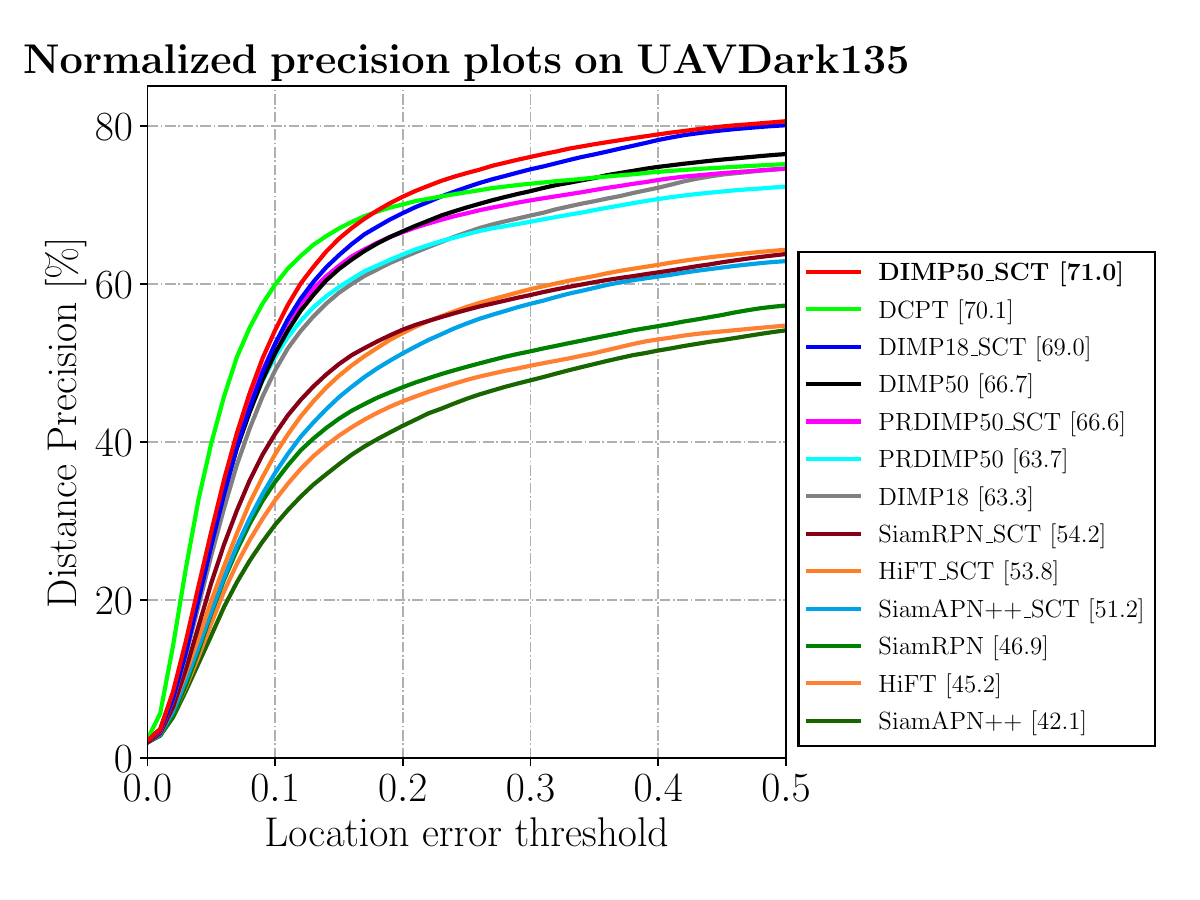}
	\vspace{-0.75cm}
\end{figure*}
\begin{figure*}[h]	
	\raggedright
	\includegraphics[width=0.325\linewidth]{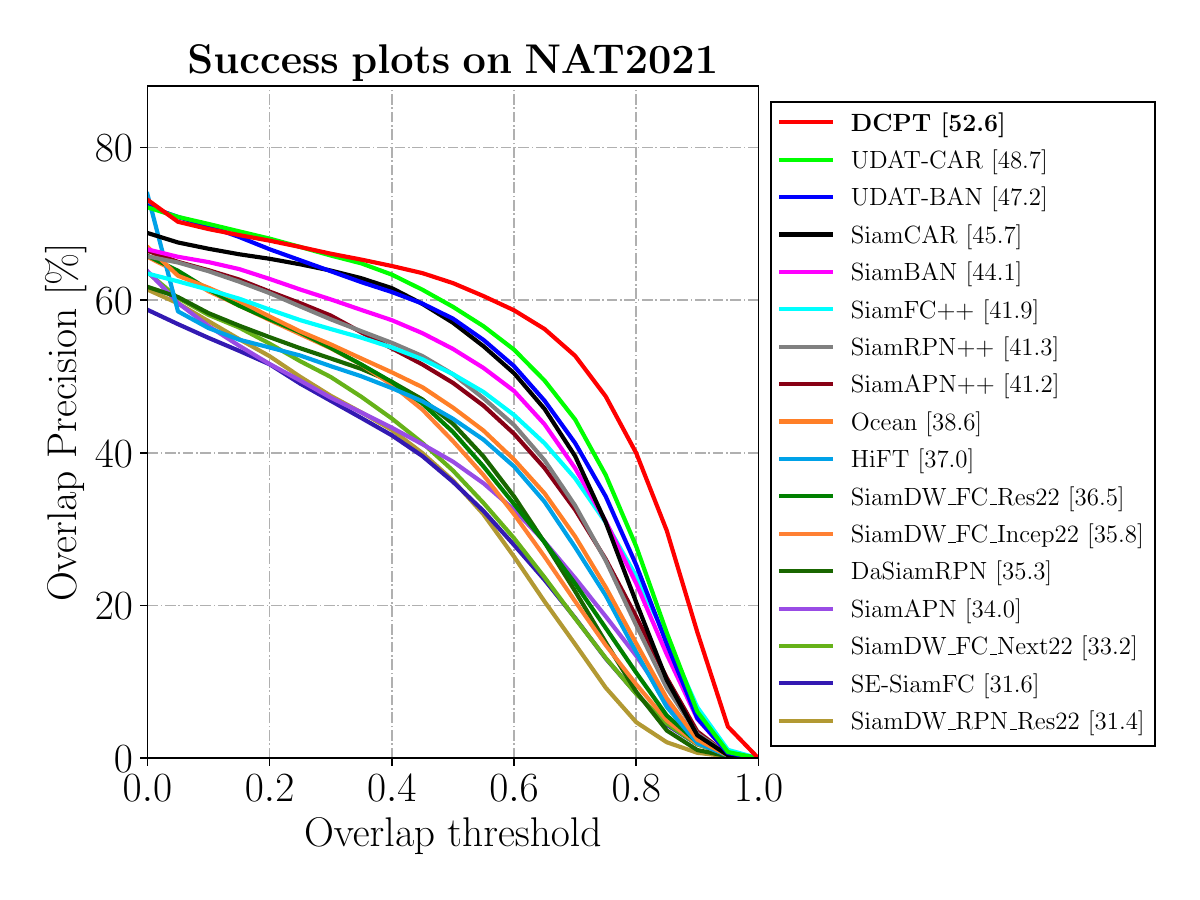}
	\includegraphics[width=0.325\linewidth]{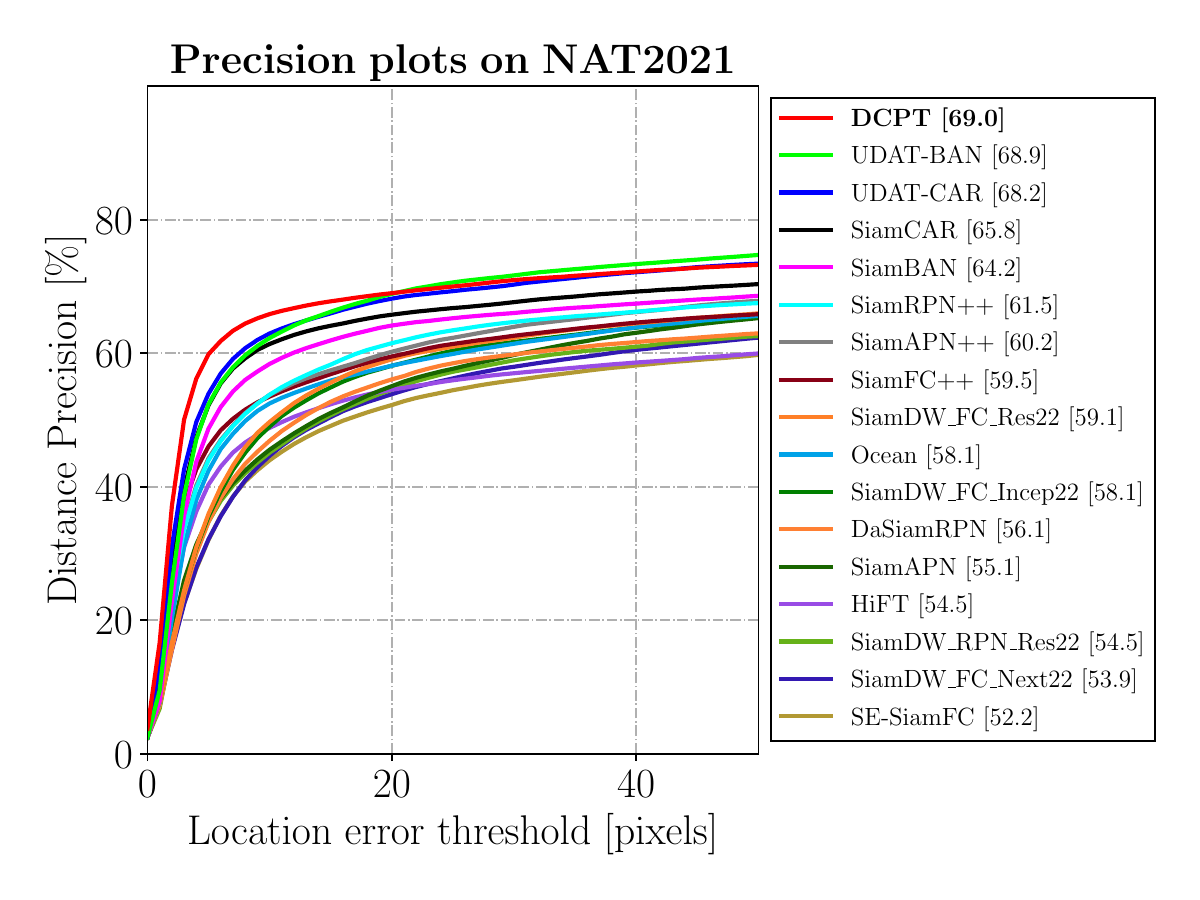}
	\includegraphics[width=0.325\linewidth]{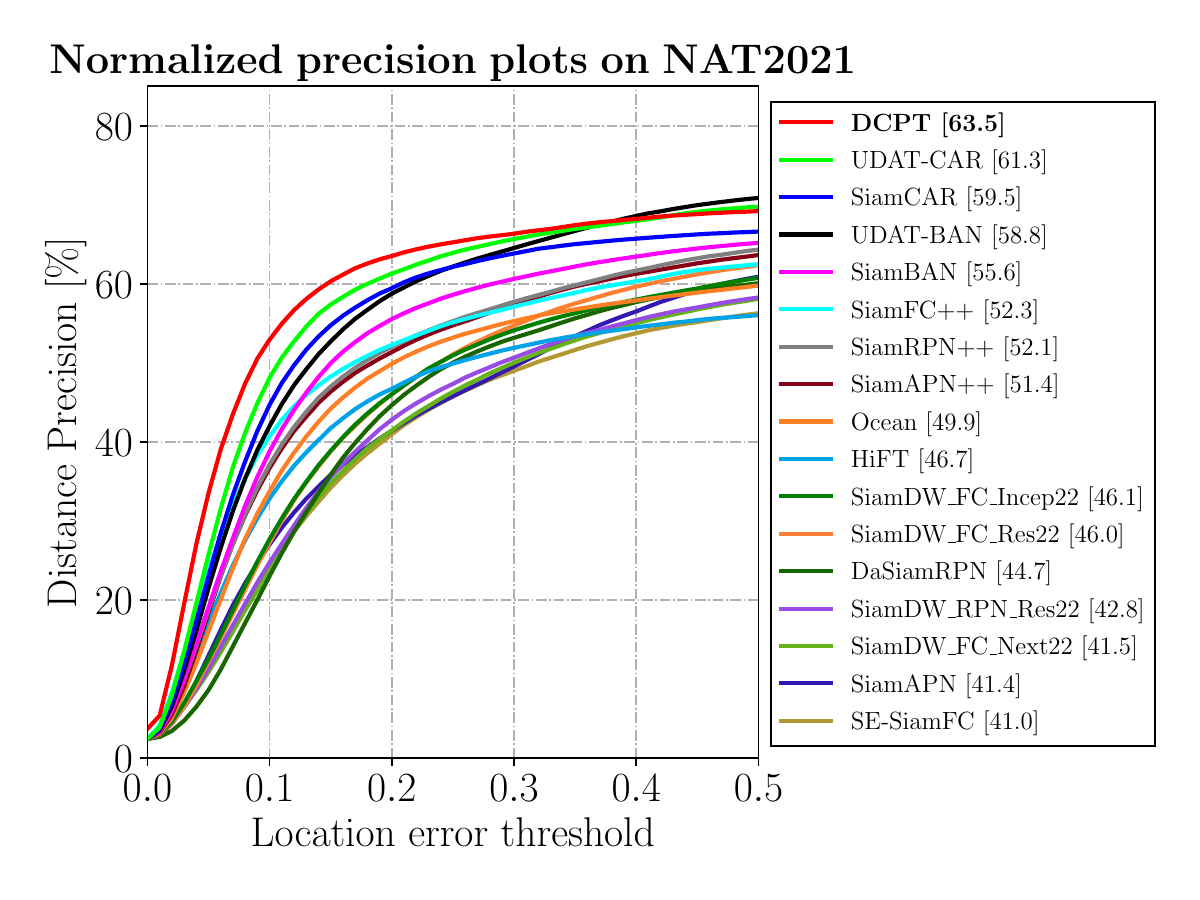}
	\vspace{-0.5cm}
	\caption
	{
		Overall performance of DCPT and other SOTA trackers on UAVDark135 \cite{UAVDark135} (the first row) and NAT2021 \cite{nat2021} (the second row) benchmarks. 
	}
	\label{fig:all}
	\vspace{-0.5cm}
\end{figure*}

\begin{table*}[!t]
	\renewcommand\arraystretch{0.9}
	\centering
	\setlength{\tabcolsep}{2.5mm}
	\fontsize{6.5}{8}\selectfont
	\begin{threeparttable}
		\caption{results of the selected top 8 trackers on  DarkTrack2021~\cite{darktrack}. \textbf{\textcolor[rgb]{ 1,  0,  0}{Red}}, \textbf{\textcolor[rgb]{ 0,  1,  0}{green}}, and \textbf{\textcolor[rgb]{ 0,  0,  1}{blue}} DENOTE 1ST, 2ND, AND 3RD PLACE.}
		\vspace{0.20cm}
		\begin{tabular}{cccccccccccc}
			\toprule[0.8pt]
			Tracker & \textbf{DCPT} & DIMP50-SCT\cite{darktrack}& DIMP18\cite{bhatICCV2019} & PRDIMP50\cite{danelljanCVPR2020} & SiamRPN\cite{li2018high} & HiFT\cite{cao2021hift}&SiamAPN++-SCT \cite{darktrack} & SiamAPN++ \cite{siamapn++} \\
			\midrule
			Success & \textbf{\textcolor{red}{0.540}}& \textbf{\textcolor{blue}{0.521}} & \textbf{\textcolor{green}{0.471}}& 0.464& 0.387 & 0.374 & 0.408 & 0.377\\
			Precision & \textbf{\textcolor{green}{0.667}}&\textbf{\textcolor{red}{0.677}} & \textbf{\textcolor{blue}{0.620}}&0.580& 0.509 & 0.503& 0.537&0.489\\
			Normed Precision &\textbf{\textcolor{red}{0.646}} & \textbf{\textcolor{green}{0.633}} & \textbf{\textcolor{blue}{0.589}}&0.559 & 0.485 & 0.471& 0.511& 0.461\\
			\bottomrule[0.8pt]			
		\end{tabular}\label{tab:darktrack}
	\end{threeparttable}
	\vspace{-0.5cm}
\end{table*}

\begin{table*}[!t]
	\renewcommand\arraystretch{0.9}
	\centering
	\setlength{\tabcolsep}{1.2mm}
	\fontsize{6.5}{8}\selectfont
	\begin{threeparttable}
		\caption{results of the selected top 10 trackers on  NAT2021-L~\cite{nat2021}.
			\textbf{\textcolor[rgb]{ 1,  0,  0}{Red}}, \textbf{\textcolor[rgb]{ 0,  1,  0}{green}}, and \textbf{\textcolor[rgb]{ 0,  0,  1}{blue}} DENOTE 1ST, 2ND, AND 3RD PLACE.
		}
		\vspace{0.20cm}
		\begin{tabular}{cccccccccccc}
			\toprule[0.8pt]
			Tracker & \textbf{DCPT} & UDAT-CAR\cite{nat2021}& UDAT-BAN\cite{nat2021} & SiamRPN++\cite{li2019siamrpn++} & SiamFC++\cite{xu2020siamfc++}&SiamAPN++ \cite{siamapn++} & SiamAPN \cite{siamapn} & Ocean \cite{zhang2020ocean} & HiFT \cite{cao2021hift} & DaSiamRPN \cite{zhu2018dasiamrpn}\\
			\midrule
			Success & \textbf{\textcolor{red}{0.474}}& \textbf{\textcolor{blue}{0.378}} & \textbf{\textcolor{green}{0.353}}& 0.300& 0.298 & 0.280 & 0.242 & 0.316& 0.288 & 0.273\\
			Precision & \textbf{\textcolor{red}{0.599}}&\textbf{\textcolor{green}{0.504}} & \textbf{\textcolor{blue}{0.494}}&0.429& 0.423 & 0.400& 0.377&0.451&0.430&0.429\\
			Normed Precision & \textbf{\textcolor{red}{0.546}} & \textbf{\textcolor{green}{0.447}} & \textbf{\textcolor{blue}{0.437}}&0.358 & 0.356 & 0.327& 0.277& 0.400& 0.330& 0.337\\
			\bottomrule[0.8pt]			
		\end{tabular}\label{tab:nat21}
	\end{threeparttable}
	\vspace{-0.5cm}
\end{table*}

\textbf{Prompt Tuning.}
In this stage, the foundation model is frozen, and only prompt-related parameters are tuned. 
We construct three nighttime tracking datasets for prompt learning, \ie BDD100K-Night, SHIFT-Night and ExDark~\cite{ExDark}.
In particular, we pick the images in BDD100K~\cite{bdd100k} and SHIFT~\cite{shift} with the label “night” inside to build BDD100K-Night and SHIFT-Night.
We tune the prompt modules for 60 epochs and the initial learning rate is set to $4\times 10^{-4}$ and decreased by the factor of 10 after 48 epochs. Other settings are the same as foundation model.

\subsection{Overall Performance}

\begin{figure}[t]
	\centering
	\includegraphics[width=0.475\textwidth]{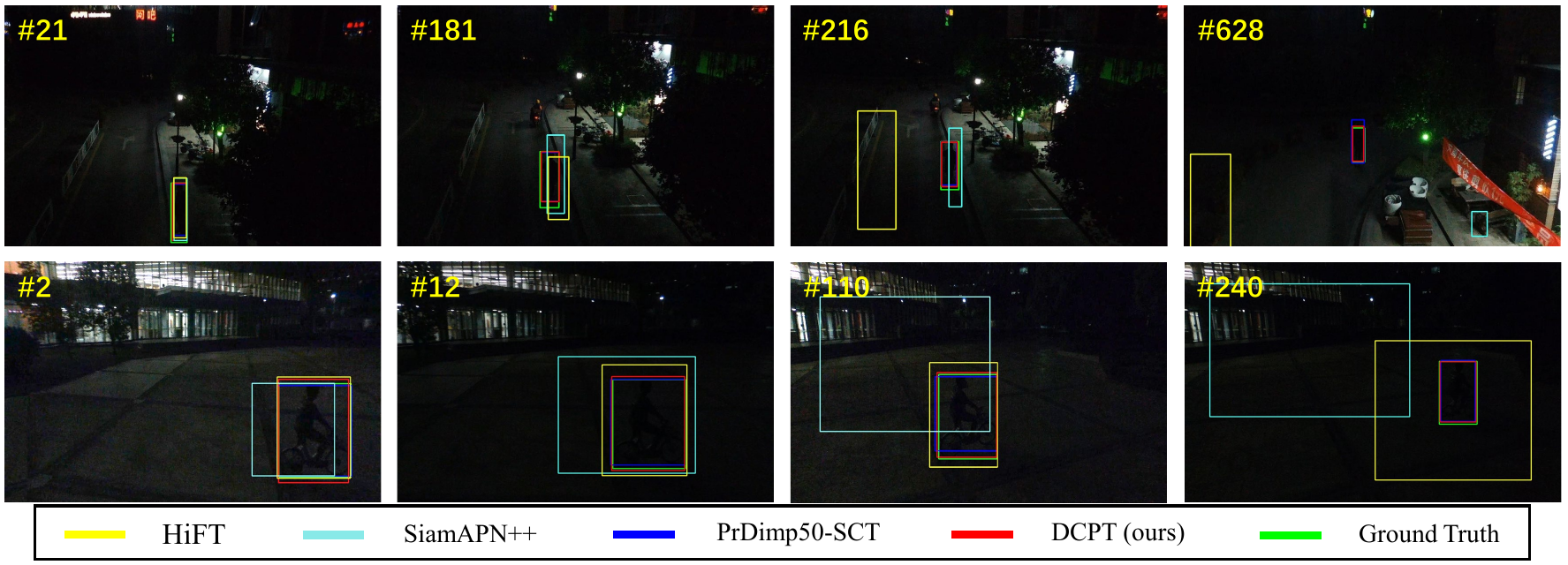}
	\vspace{-3mm}
	\caption{
		Visualization of tracking 
		in representative nighttime scenarios.}
	\label{fig:vis}
	\vspace{-3.mm}
\end{figure}

\textbf{UAVDark135 and DarkTrack2021.}
UAVDark135~\cite{UAVDark135} and DarkTrack2021~\cite{darktrack} are two of the most commonly used benchmarks for nighttime tracking. Fig.~\ref{fig:all} shows the success, precision and normalized precision curve for UAVDark135 and Tab.~\ref{tab:darktrack} shows the result for DarkTrack2021. Our method is superior than other SOTA trackers and achieves a success score of 57.7\% and 54.0\% in these two benchmarks respectively, which beats the second best trackers by 
1.5\% and 1.9\%.
We also pick several representative frames from 
UAVDark135 for visualization in Fig.~\ref{fig:vis}.
As we can see, DCPT can track target objects more steadily than others.

\textbf{NAT2021 and NAT2021-L.}
NAT2021~\cite{nat2021} is a 
benchmark with 12 different attributes like full-occlusion and low-ambient intensity. Therefore, it is more difficult 
for accurate tracking. 
Despite being challenging,
our tracker shows remarkable results as depicted in the second row of Fig.~\ref{fig:all}. DCPT ranks first in terms of all success, normed precision and precision scores.
NAT2021-L~\cite{nat2021} is a long-term tracking benchmark which involves multiple challenging attributes and more than 1400 frames in each sequence. As shown in Tab.~\ref{tab:nat21}, DCPT shows surprising results, outperforming previous SOTA trackers UDAT-CAR and UDAT-BAN by almost 10 percent and achieving 47.4\% in terms of success score and 59.9\% in terms of precision score.

\subsection{Attribute-based Analysis}
\begin{figure}[!t]
	\centering
	\includegraphics[trim=-18 0 0 0, clip, width=0.425\textwidth]{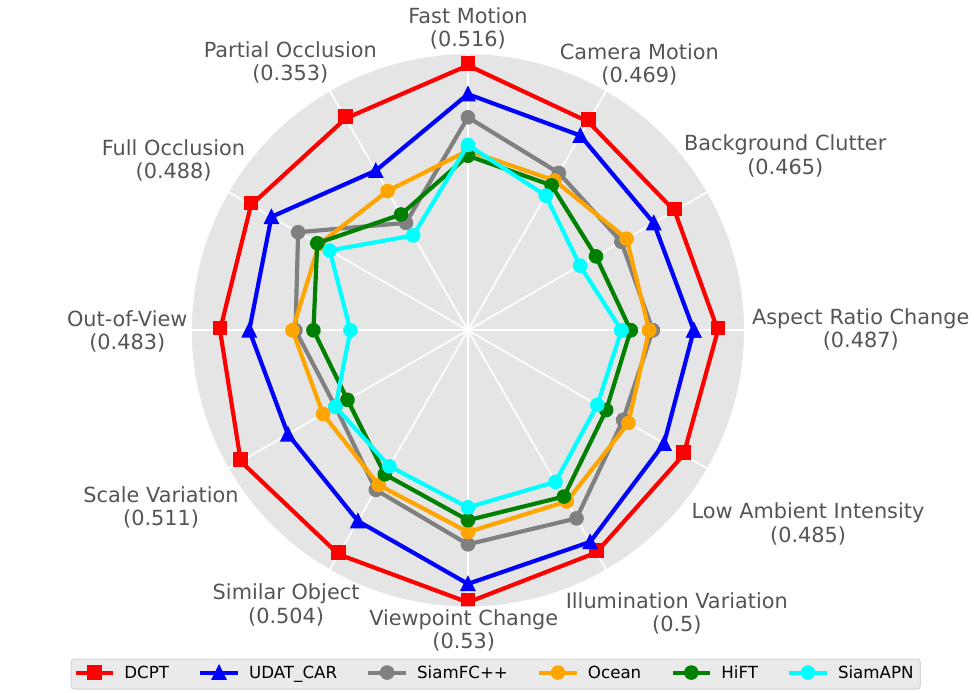}
	\vspace{-3mm}
	\caption{Success score comparison of different attributes on NAT2021~\cite{nat2021}.}
	\label{fig:att_ladar}
	\vspace{-3mm}
\end{figure}

\begin{figure}[t]	
	\raggedright
	\includegraphics[width=0.475\linewidth]{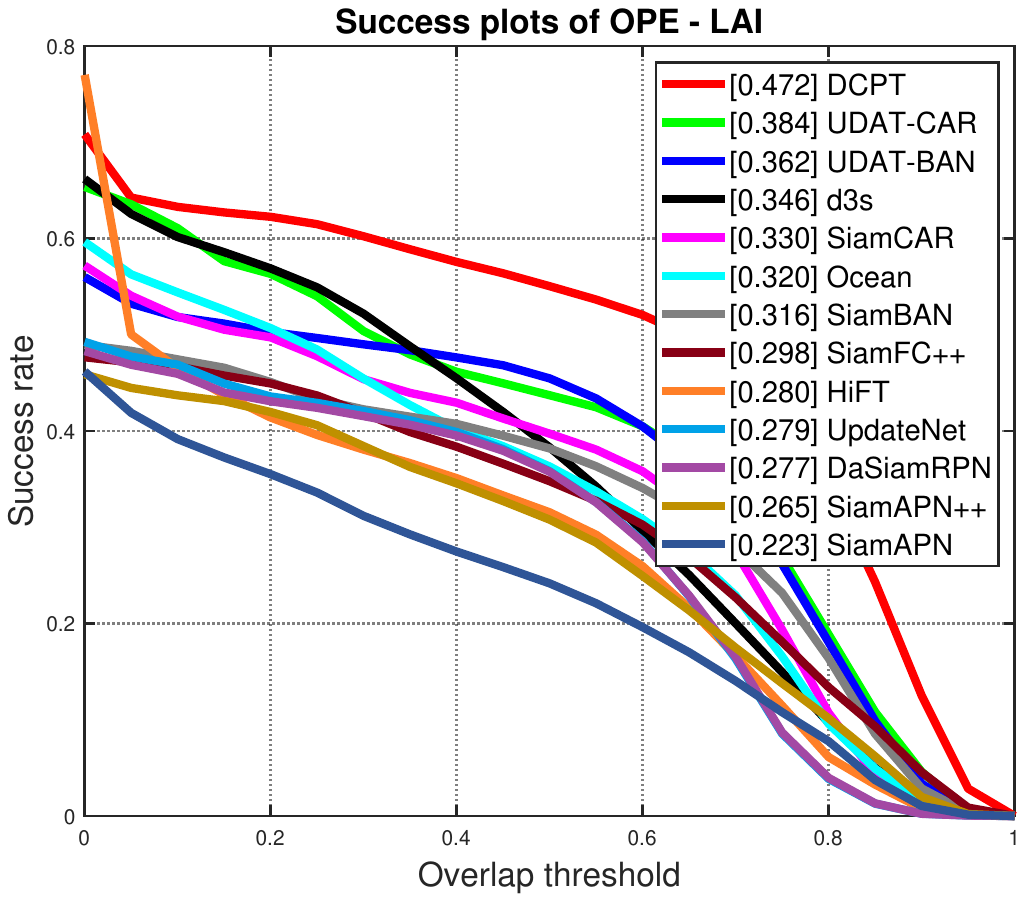}
	\includegraphics[width=0.475\linewidth]{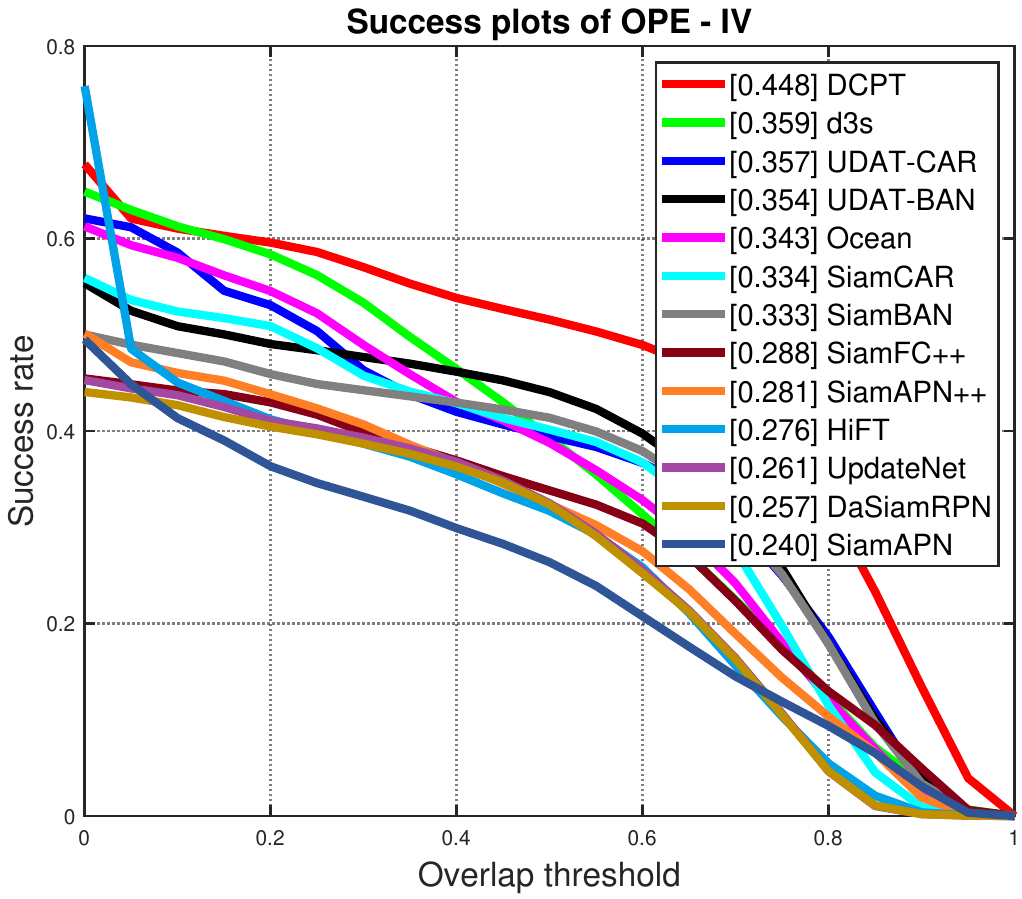}
	\vspace{-4mm}
	\caption
	{
		Success plots of illumination-related attributes on NAT2021-L~\cite{nat2021}.
	}
	\vspace{-2mm}
	\label{fig:att_pr}
\end{figure}

The superiority of DCPT is further validated by attribute-based comparison in NAT2021~\cite{nat2021}. This dataset contains twelve attributes, \eg aspect ratio change, background clutter, camera motion, \etc 
Results of success score are provided in Fig.~\ref{fig:att_ladar}. 
DCPT achieves the best scores in all 12 scenarios. Especially in scale variation, viewpoint change and illumination variation, 
DCPT achieves impressive performance which mainly benefits from 
critical and effective darkness clue prompts in nighttime circumstances.
Besides, 
we also report the result plots of illumination-related attributes on NAT2021-L~\cite{nat2021}.
As shown in Fig.\ref{fig:att_pr}, our tracker 
leads dramatically in these two attributes and thereby proves the effectiveness of our darkness clue prompt tracking paradigm. 


\subsection{Ablation Studies}
To verify the effectiveness of these proposed components, we will gradually introduce them into the base tracker in the subsequent subsection. We will then provide corresponding results using UAVDark135~\cite{UAVDark135} and NAT2021~\cite{nat2021}.

\textbf{Base+DCP.}
Darkness clue prompter (DCP) is the core component of our tracker. 
It utilizes a back-projection block to iteratively mine critical darkness clue prompts for foundation daytime tracker to enable the ability to see more clearly in the dark.
As shown in Tab.\ref{tab:power_of_mm}, DCP boosts the base tracker with improvements of 1.95\% and 2.97\% success scores on UAVDark135 and DarkTrack2021, respectively. 

\textbf{Base+DCP+GFA\_pp.} 
We add the gated feature aggregated for adjacent prompts as described in Sec.~\ref{sec:gfa}.
As illustrated in Tab.~\ref{tab:power_of_mm}, on DarkTrack2021, 
the tracker obtains a success score of 53.44\%, which is 4.39\% higher than the foundation tracker.
On UAVDark135, the performance also boosts to 57.51\% in terms of success score.
The results indicate that performing gated fusion facilitates the fusion of darkness prompts at different semantic hierarchies.

\textbf{Base+DCP+GFA\_{pp,pb}.}
Further, we continue to add gated feature aggregation between prompts and foundation features.
As reported in Tab.~\ref{tab:power_of_mm}, remarkable performance gains are consistently obtained.
Gated feature aggregation here allows adaptive injection of prompts into different foundation tokens. 
Ultimately, the obtained performance has a tremendous improvement over the daytime foundation tracker.
With only 3.03M (3.3\%) trainable parameters, DCPT improves on UAVDark135 by 2.67\% and 4.03\% in terms of success and precision scores. On DarkTrack2021, the improvements even come to 4.93\% and 6.64\%.

\subsection{Real-World Testing}
\begin{figure}[!t]
	\centering
	\includegraphics[width=0.46\textwidth]{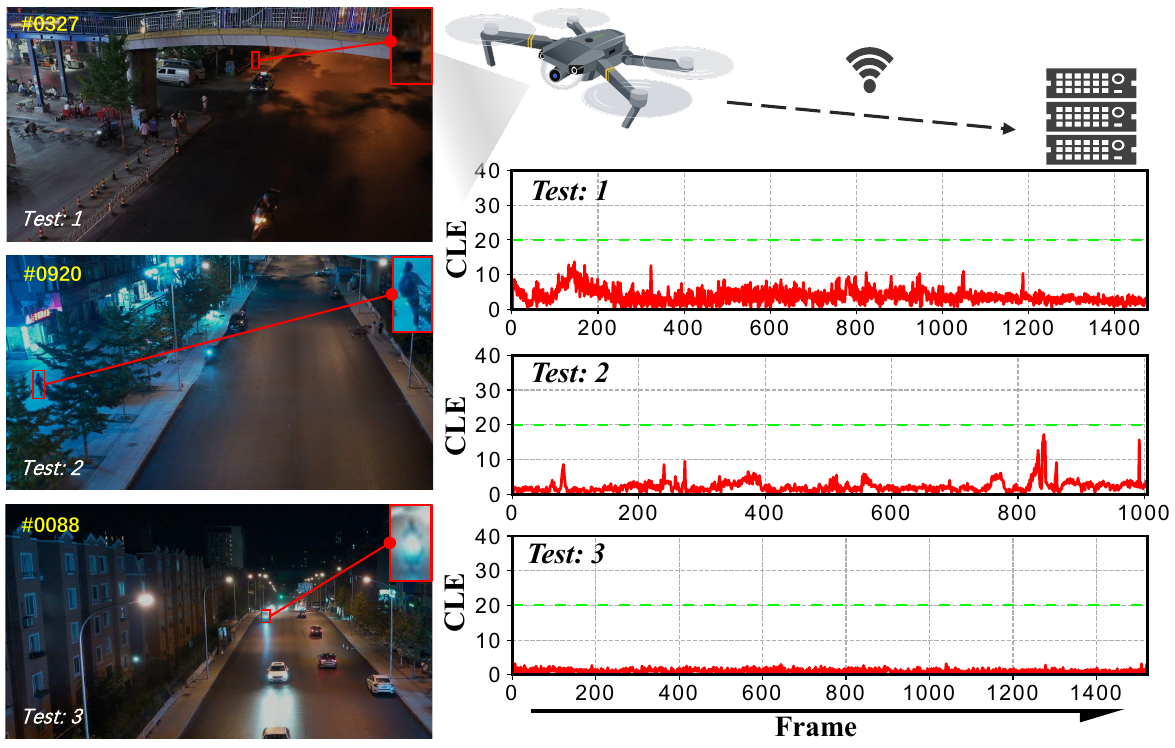}
	\vspace{-3mm}
	\caption{The real-world nighttime UAV tracking testing. The frame-wise performance is presented in terms of CLE plots. The errors below the green dashed lines (CLE$=$20 pixels) are usually considered acceptable.}
	\label{fig:realworld}
	\vspace{-3mm}
\end{figure}

We perform a series of real-world tests to further verify the feasibility and generalization of DCPT.
The on-board camera on the UAV captures nighttime scenes and transmits the captured images to the workstation in real time through Wi-Fi communication.
The workstation is a computer with an Intel(R) i7-9700K CPU @3.60GHz and an Nvidia 2080ti GPU, which can process the received images 
with a promising speed of over 30 fps/720P.
As shown in Fig.~\ref{fig:realworld}, the main challenges  are low resolution, partial occlusion, and low ambient intensity, yet DCPT achieves favorable performance with the average CLE (center location error) of 3.81, 2.19, and 1.04 pixels, and all the test frames have a CLE of less than 20 pixels.
The real-world testing demonstrates the feasibility of the DCPT paradigm by injecting learned darkness clue prompts into the daytime tracker to significantly improve its tracking performance in complex nighttime circumstances.

\begin{table}[ht]
	\renewcommand\arraystretch{1.2}
	\vspace{-0.2cm}
	\centering
	\caption{Ablation studies on multiple nighttime benchmarks.
		Params$^\dag$ denotes the number of trainable parameters.
	}
	\vspace{-3mm}
	\fontsize{8}{10}\selectfont  
	\setlength{\tabcolsep}{0.6mm}{
		\resizebox{\linewidth}{!}{%
			\begin{tabular}{c|c|cc|cc}
				\hline
				\multirow{2}{*}{Method} &
				\multirow{2}{*}{Params$^{\dag}$} &
				\multicolumn{2}{c|}{UAVDark135~\cite{UAVDark135}} &
				\multicolumn{2}{c}{DarkTrack2021~\cite{darktrack}}  \\
				\cline{3-6}
				&& Success($\uparrow$) & Precision($\uparrow$) & Success($\uparrow$) & Precision($\uparrow$)  \\
				\hline
				{Base} &89.96M & 55.07 & 66.23 & 49.05 & 60.06 \\
				
				{Base+DCP} &3.03M (3.3\%)&57.02 (1.95$\uparrow$) &69.23&52.02 (2.97$\uparrow$) &64.39  \\
				
				{Base+DCP+GFA\_pp} &3.03M (3.3\%)& 57.51 (2.44$\uparrow$)  & 69.35 & 53.44 (4.39$\uparrow$) &65.99  \\
				
				{Base+DCP+GFA\_pp,pb} &3.03M (3.3\%)&\textbf{\textcolor{red}{57.74}} (2.67$\uparrow$)  &\textbf{\textcolor{red}{70.26}} &\textbf{\textcolor{red}{53.98}} (4.93$\uparrow$) &\textbf{\textcolor{red}{66.70}}  \\

				\hline
	\end{tabular}}}
	\vspace{-4mm}
	\label{tab:power_of_mm}
\end{table}

\section{Conclusion}
This work proposes DCPT, a new end-to-end framework for nighttime UAV tracking. DCPT learns to generate darkness clue prompts that stimulate the tracking capabilities of a fixed daytime tracker for nighttime operation. The proposed Darkness Clue Prompter mines crucial darkness cues, while the gated aggregation mechanism enables adaptive fusion of prompts across layers and with tracker features.
Compared to prior methods, DCPT inherits robust tracking from a daytime model trained on massive datasets, in a streamlined and end-to-end trainable architecture. Extensive experiments validate its state-of-the-art effectiveness for nighttime tracking.

\section*{Acknowledgment}
The work is supported by the National Natural Science Foundation of China (No.62206040, 62293540, 62293542).

	\bibliographystyle{IEEEtran}
	\normalem
	\bibliography{icra}

\begin{thebibliography}{10}
\providecommand{\url}[1]{#1}
\csname url@samestyle\endcsname
\providecommand{\newblock}{\relax}
\providecommand{\bibinfo}[2]{#2}
\providecommand{\BIBentrySTDinterwordspacing}{\spaceskip=0pt\relax}
\providecommand{\BIBentryALTinterwordstretchfactor}{4}
\providecommand{\BIBentryALTinterwordspacing}{\spaceskip=\fontdimen2\font plus
\BIBentryALTinterwordstretchfactor\fontdimen3\font minus
  \fontdimen4\font\relax}
\providecommand{\BIBforeignlanguage}[2]{{%
\expandafter\ifx\csname l@#1\endcsname\relax
\typeout{** WARNING: IEEEtran.bst: No hyphenation pattern has been}%
\typeout{** loaded for the language `#1'. Using the pattern for}%
\typeout{** the default language instead.}%
\else
\language=\csname l@#1\endcsname
\fi
#2}}
\providecommand{\BIBdecl}{\relax}
\BIBdecl

\bibitem{tian2011video}
B.~Tian, Q.~Yao, Y.~Gu, K.~Wang, and Y.~Li, ``Video processing techniques for
  traffic flow monitoring: A survey,'' in \emph{2011 14th international IEEE
  conference on intelligent transportation systems (ITSC)}.\hskip 1em plus
  0.5em minus 0.4em\relax IEEE, 2011, pp. 1103--1108.

\bibitem{bonatti2019towards}
R.~Bonatti, C.~Ho, W.~Wang, S.~Choudhury, and S.~Scherer, ``Towards a robust
  aerial cinematography platform: Localizing and tracking moving targets in
  unstructured environments,'' in \emph{2019 IEEE/RSJ International Conference
  on Intelligent Robots and Systems (IROS)}.\hskip 1em plus 0.5em minus
  0.4em\relax IEEE, 2019, pp. 229--236.

\bibitem{al2019appearance}
A.~Al-Kaff, M.~J. G{\'o}mez-Silva, F.~M. Moreno, A.~De~La~Escalera, and J.~M.
  Armingol, ``An appearance-based tracking algorithm for aerial search and
  rescue purposes,'' \emph{Sensors}, vol.~19, no.~3, p. 652, 2019.

\bibitem{siamapn++}
Z.~Cao, C.~Fu, J.~Ye, B.~Li, and Y.~Li, ``Siamapn++: Siamese attentional
  aggregation network for real-time uav tracking,'' in \emph{2021 IEEE/RSJ
  International Conference on Intelligent Robots and Systems (IROS)}.\hskip 1em
  plus 0.5em minus 0.4em\relax IEEE, 2021, pp. 3086--3092.

\bibitem{transt}
X.~Chen, B.~Yan, J.~Zhu, D.~Wang, X.~Yang, and H.~Lu, ``Transformer tracking,''
  in \emph{CVPR}, 2021.

\bibitem{ostrack}
B.~Ye, H.~Chang, B.~Ma, S.~Shan, and X.~Chen, ``Joint feature learning and
  relation modeling for tracking: A one-stream framework,'' in
  \emph{ECCV}.\hskip 1em plus 0.5em minus 0.4em\relax Springer, 2022, pp.
  341--357.

\bibitem{alexnet}
A.~Krizhevsky, I.~Sutskever, and G.~E. Hinton, ``Imagenet classification with
  deep convolutional neural networks,'' in \emph{NIPS}, 2012.

\bibitem{resnet}
K.~He, X.~Zhang, S.~Ren, and J.~Sun, ``Deep residual learning for image
  recognition,'' in \emph{CVPR}, 2016.

\bibitem{vit}
A.~Dosovitskiy, L.~Beyer, A.~Kolesnikov, D.~Weissenborn, X.~Zhai,
  T.~Unterthiner, M.~Dehghani, M.~Minderer, G.~Heigold, S.~Gelly, J.~Uszkoreit,
  and N.~Houlsby, ``An image is worth 16x16 words: Transformers for image
  recognition at scale,'' \emph{ICLR}, 2021.

\bibitem{uav}
M.~Mueller, N.~Smith, and B.~Ghanem, ``A benchmark and simulator for {UAV}
  tracking,'' in \emph{ECCV}, 2016.

\bibitem{got10k}
L.~Huang, X.~Zhao, and K.~Huang, ``Got-10k: A large high-diversity benchmark
  for generic object tracking in the wild,'' \emph{TPAMI}, 2019.

\bibitem{trackingnet}
M.~Muller, A.~Bibi, S.~Giancola, S.~Alsubaihi, and B.~Ghanem, ``Tracking{N}et:
  A large-scale dataset and benchmark for object tracking in the wild,'' in
  \emph{ECCV}, 2018.

\bibitem{lin2014multi}
H.~Lin and Z.~Shi, ``Multi-scale retinex improvement for nighttime image
  enhancement,'' \emph{optik}, vol. 125, no.~24, pp. 7143--7148, 2014.

\bibitem{highlightnet}
C.~Fu, H.~Dong, J.~Ye, G.~Zheng, S.~Li, and J.~Zhao, ``Highlightnet:
  Highlighting low-light potential features for real-time uav tracking,'' in
  \emph{2022 IEEE/RSJ International Conference on Intelligent Robots and
  Systems (IROS)}.\hskip 1em plus 0.5em minus 0.4em\relax IEEE, 2022, pp.
  12\,146--12\,153.

\bibitem{darklighter}
J.~Ye, C.~Fu, G.~Zheng, Z.~Cao, and B.~Li, ``Darklighter: Light up the darkness
  for uav tracking,'' in \emph{2021 IEEE/RSJ International Conference on
  Intelligent Robots and Systems (IROS)}.\hskip 1em plus 0.5em minus
  0.4em\relax IEEE, 2021, pp. 3079--3085.

\bibitem{adtrack}
B.~Li, C.~Fu, F.~Ding, J.~Ye, and F.~Lin, ``Adtrack: Target-aware dual filter
  learning for real-time anti-dark uav tracking,'' in \emph{2021 IEEE
  international conference on robotics and automation (ICRA)}.\hskip 1em plus
  0.5em minus 0.4em\relax IEEE, 2021, pp. 496--502.

\bibitem{nat2021}
J.~Ye, C.~Fu, G.~Zheng, D.~P. Paudel, and G.~Chen, ``Unsupervised domain
  adaptation for nighttime aerial tracking,'' in \emph{Proceedings of the
  IEEE/CVF Conference on Computer Vision and Pattern Recognition}, 2022, pp.
  8896--8905.

\bibitem{doprompt}
Z.~Zheng, X.~Yue, K.~Wang, and Y.~You, ``Prompt vision transformer for domain
  generalization,'' \emph{arXiv preprint arXiv:2208.08914}, 2022.

\bibitem{vpt}
M.~Jia, L.~Tang, B.-C. Chen, C.~Cardie, S.~Belongie, B.~Hariharan, and S.-N.
  Lim, ``Visual prompt tuning,'' in \emph{ECCV}, 2022.

\bibitem{vipt}
J.~Zhu, S.~Lai, X.~Chen, D.~Wang, and H.~Lu, ``Visual prompt multi-modal
  tracking,'' in \emph{Proceedings of the IEEE/CVF Conference on Computer
  Vision and Pattern Recognition}, 2023, pp. 9516--9526.

\bibitem{dbpn}
M.~Haris, G.~Shakhnarovich, and N.~Ukita, ``Deep back-projection networks for
  super-resolution,'' in \emph{Proceedings of the IEEE conference on computer
  vision and pattern recognition}, 2018, pp. 1664--1673.

\bibitem{siameserpn}
B.~Li, J.~Yan, W.~Wu, Z.~Zhu, and X.~Hu, ``High performance visual tracking
  with siamese region proposal network,'' in \emph{CVPR}, 2018.

\bibitem{dimp}
G.~Bhat, M.~Danelljan, L.~V. Gool, and R.~Timofte, ``Learning discriminative
  model prediction for tracking,'' in \emph{ICCV}, 2019.

\bibitem{siamapn}
C.~Fu, Z.~Cao, Y.~Li, J.~Ye, and C.~Feng, ``Siamese anchor proposal network for
  high-speed aerial tracking,'' in \emph{2021 IEEE International Conference on
  Robotics and Automation (ICRA)}.\hskip 1em plus 0.5em minus 0.4em\relax IEEE,
  2021, pp. 510--516.

\bibitem{darktrack}
J.~Ye, C.~Fu, Z.~Cao, S.~An, G.~Zheng, and B.~Li, ``Tracker meets night: A
  transformer enhancer for uav tracking,'' \emph{IEEE Robotics and Automation
  Letters}, vol.~7, no.~2, pp. 3866--3873, 2022.

\bibitem{bahng2022visual}
H.~Bahng, A.~Jahanian, S.~Sankaranarayanan, and P.~Isola, ``Visual prompting:
  Modifying pixel space to adapt pre-trained models,'' \emph{arXiv preprint
  arXiv:2203.17274}, 2022.

\bibitem{lasot}
H.~Fan, L.~Lin, F.~Yang, P.~Chu, G.~Deng, S.~Yu, H.~Bai, Y.~Xu, C.~Liao, and
  H.~Ling, ``{LaSOT}: A high-quality benchmark for large-scale single object
  tracking,'' in \emph{CVPR}, 2019.

\bibitem{li2019siamrpn++}
B.~Li, W.~Wu, Q.~Wang, F.~Zhang, J.~Xing, and J.~Yan, ``Siamrpn++: Evolution of
  siamese visual tracking with very deep networks,'' in \emph{Proceedings of
  the IEEE/CVF conference on computer vision and pattern recognition}, 2019,
  pp. 4282--4291.

\bibitem{irani1991improving}
M.~Irani and S.~Peleg, ``Improving resolution by image registration,''
  \emph{CVGIP: Graphical models and image processing}, vol.~53, no.~3, pp.
  231--239, 1991.

\bibitem{wang2020lightening}
L.-W. Wang, Z.-S. Liu, W.-C. Siu, and D.~P. Lun, ``Lightening network for
  low-light image enhancement,'' \emph{IEEE Transactions on Image Processing},
  vol.~29, pp. 7984--7996, 2020.

\bibitem{yoo2023improving}
S.~Yoo, E.~Kim, D.~Jung, J.~Lee, and S.~Yoon, ``Improving visual prompt tuning
  for self-supervised vision transformers,'' \emph{arXiv preprint
  arXiv:2306.05067}, 2023.

\bibitem{giou}
H.~Rezatofighi, N.~Tsoi, J.~Gwak, A.~Sadeghian, I.~D. Reid, and S.~Savarese,
  ``Generalized intersection over union: {A} metric and a loss for bounding box
  regression,'' in \emph{CVPR}, 2019.

\bibitem{coco}
T.-Y. Lin, M.~Maire, S.~J. Belongie, L.~D. Bourdev, R.~B. Girshick, J.~Hays,
  P.~Perona, D.~Ramanan, P.~Doll{\'a}r, and C.~L. Zitnick, ``{Microsoft COCO}:
  Common objects in context,'' in \emph{ECCV}, 2014.

\bibitem{mae}
K.~He, X.~Chen, S.~Xie, Y.~Li, P.~Doll{\'a}r, and R.~Girshick, ``Masked
  autoencoders are scalable vision learners,'' in \emph{CVPR}, 2022, pp.
  16\,000--16\,009.

\bibitem{adamw}
I.~Loshchilov and F.~Hutter, ``Decoupled weight decay regularization,'' in
  \emph{ICLR}, 2018.

\bibitem{UAVDark135}
B.~Li, C.~Fu, F.~Ding, J.~Ye, and F.~Lin, ``All-day object tracking for
  unmanned aerial vehicle,'' \emph{IEEE Transactions on Mobile Computing},
  2022.

\bibitem{bhatICCV2019}
G.~Bhat, M.~Danelljan, L.~V. Gool, and R.~Timofte, ``Learning discriminative
  model prediction for tracking,'' in \emph{Proceedings of the IEEE/CVF
  international conference on computer vision}, 2019, pp. 6182--6191.

\bibitem{danelljanCVPR2020}
M.~Danelljan, L.~V. Gool, and R.~Timofte, ``Probabilistic regression for visual
  tracking,'' in \emph{Proceedings of the IEEE/CVF conference on computer
  vision and pattern recognition}, 2020, pp. 7183--7192.

\bibitem{li2018high}
B.~Li, J.~Yan, W.~Wu, Z.~Zhu, and X.~Hu, ``High performance visual tracking
  with siamese region proposal network,'' in \emph{Proceedings of the IEEE
  conference on computer vision and pattern recognition}, 2018, pp. 8971--8980.

\bibitem{cao2021hift}
Z.~Cao, C.~Fu, J.~Ye, B.~Li, and Y.~Li, ``Hift: Hierarchical feature
  transformer for aerial tracking,'' in \emph{Proceedings of the IEEE/CVF
  International Conference on Computer Vision}, 2021, pp. 15\,457--15\,466.

\bibitem{xu2020siamfc++}
Y.~Xu, Z.~Wang, Z.~Li, Y.~Yuan, and G.~Yu, ``Siamfc++: Towards robust and
  accurate visual tracking with target estimation guidelines,'' in
  \emph{Proceedings of the AAAI conference on artificial intelligence},
  vol.~34, no.~07, 2020, pp. 12\,549--12\,556.

\bibitem{zhang2020ocean}
Z.~Zhang, H.~Peng, J.~Fu, B.~Li, and W.~Hu, ``Ocean: Object-aware anchor-free
  tracking,'' in \emph{Computer Vision--ECCV 2020: 16th European Conference,
  Glasgow, UK, August 23--28, 2020, Proceedings, Part XXI 16}.\hskip 1em plus
  0.5em minus 0.4em\relax Springer, 2020, pp. 771--787.

\bibitem{zhu2018dasiamrpn}
Z.~Zhu, Q.~Wang, B.~Li, W.~Wu, J.~Yan, and W.~Hu, ``Distractor-aware siamese
  networks for visual object tracking,'' in \emph{Proceedings of the European
  conference on computer vision (ECCV)}, 2018, pp. 101--117.

\bibitem{ExDark}
Y.~P. Loh and C.~S. Chan, ``Getting to know low-light images with the
  exclusively dark dataset,'' \emph{Computer Vision and Image Understanding},
  vol. 178, pp. 30--42, 2019.

\bibitem{bdd100k}
F.~Yu, H.~Chen, X.~Wang, W.~Xian, Y.~Chen, F.~Liu, V.~Madhavan, and T.~Darrell,
  ``Bdd100k: A diverse driving dataset for heterogeneous multitask learning,''
  in \emph{Proceedings of the IEEE/CVF conference on computer vision and
  pattern recognition}, 2020, pp. 2636--2645.

\bibitem{shift}
T.~Sun, M.~Segu, J.~Postels, Y.~Wang, L.~Van~Gool, B.~Schiele, F.~Tombari, and
  F.~Yu, ``Shift: a synthetic driving dataset for continuous multi-task domain
  adaptation,'' in \emph{Proceedings of the IEEE/CVF Conference on Computer
  Vision and Pattern Recognition}, 2022, pp. 21\,371--21\,382.

\end{thebibliography}
	
\end{document}